\documentclass[10pt,twocolumn,letterpaper]{article}

\usepackage[pagenumbers]{cvpr} 
\usepackage{marvosym}
\usepackage{xcolor}

\definecolor{cvprblue}{rgb}{0.21,0.49,0.74}
\usepackage[pagebackref,breaklinks,colorlinks,allcolors=cvprblue]{hyperref}

\usepackage{booktabs}

\definecolor{cvprpink}{RGB}{212,31,134}

\hypersetup{
    colorlinks=true,
    linkcolor=cvprblue,
    citecolor=cvprblue,
    urlcolor=cvprpink
}

\def\paperID{20150} 
\def\confName{CVPR}
\def\confYear{2026}
\newcommand{\ours}{PhysGS}
\title{\ours{}: Bayesian-Inferred Gaussian Splatting for Physical Property Estimation}

\author{
Samarth Chopra$^{1,\;\textrm{\Letter}},$\qquad
Jing Liang$^{2},$\qquad
Gershom Seneviratne$^{1},$\qquad
Dinesh Manocha$^{1}$\\[6pt]
$^{1}$University of Maryland, College Park\qquad
$^{2}$Stanford University\\[4pt]
{\tt\small sachopra@umd.edu,\; jinglms@stanford.edu,\; gershom@umd.edu,\; dmanocha@umd.edu}
}

\begin{document}
\maketitle
\begin{abstract}

Understanding physical properties such as friction, stiffness, hardness, and material composition is essential for enabling robots to interact safely and effectively with their surroundings. However, existing 3D reconstruction methods focus on geometry and appearance and cannot infer these underlying physical properties. We present \textit{PhysGS}, a Bayesian-inferred extension of 3D Gaussian Splatting that estimates dense, per-point physical properties from visual cues and vision--language priors. We formulate property estimation as Bayesian inference over Gaussian splats, where material and property beliefs are iteratively refined as new observations arrive. PhysGS also models aleatoric and epistemic uncertainties, enabling uncertainty-aware object and scene interpretation. Across object-scale (ABO-500), indoor, and outdoor real-world datasets, PhysGS improves accuracy of the mass estimation by up to 22.8\%, reduces Shore hardness error by up to 61.2\%, and lowers kinetic friction error by up to 18.1\% compared to deterministic baselines. Our results demonstrate that PhysGS unifies 3D reconstruction, uncertainty modeling, and physical reasoning in a single, spatially continuous framework for dense physical property estimation. 
Additional results are available at \href{https://samchopra2003.github.io/physgs/}{https://samchopra2003.github.io/physgs}.

\end{abstract}    
\begin{figure}[t!]
    \centering
\includegraphics[width=0.45\textwidth]{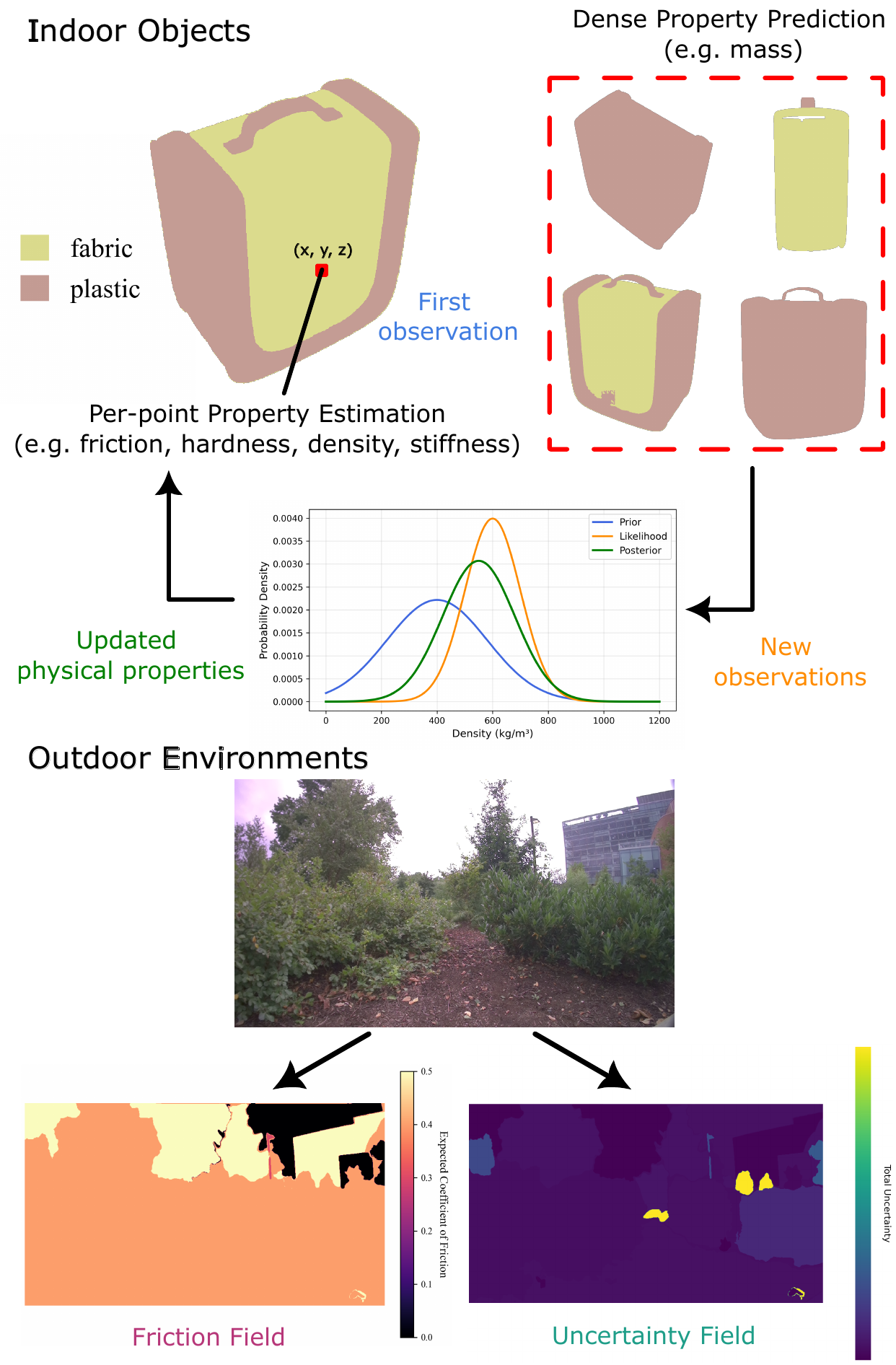}
\caption{Overview of \textit{PhysGS}. \underline{Top}: Our method estimates per-point and dense physical 
properties (e.g., friction, hardness, density, stiffness, and mass) by combining vision–language 
material priors with Bayesian updates over 3D Gaussian splats. \underline{Bottom}: \textit{PhysGS} can also be 
deployed in outdoor environments to infer scene-level properties such as friction and predictive 
uncertainty; we visualize the total uncertainty (\textit{aleatoric} + \textit{epistemic}).}
\vspace{-.5cm}
\label{fig:overview}
\end{figure}

\section{Introduction}
\label{sec:intro}

Understanding the physical properties of real-world environments is critical for enabling robots to interact safely and effectively with their surroundings~\cite{zhai2024physical, loudream, ewen2022these, chen2024identifying}. This capability is essential across a wide range of domains such as navigation~\cite{weerakoon2023graspe, liang2022adaptiveon}, manipulation~\cite{kobayashi2005physical, billard2019trends}, and surgical robotics~\cite{kobayashi2005physical, yoshizawa2005robot, omata2004real}, and is required in both complex indoor~\cite{zhai2024physical, xu2025gaussianproperty, park2025understanding} and outdoor~\cite{liang2022adaptiveon, ewen2022these, seneviratne2024cross} environments. In particular, accurately estimating physical attributes such as friction~\cite{liang2022adaptiveon, ewen2022these}, elasticity~\cite{weerakoon2023graspe}, hardness~\cite{chen2024identifying}, and density~\cite{xu2025gaussianproperty} is essential for safe and robust robot interaction with diverse and unstructured real-world scenarios.

In general, estimating physical properties from visual sensors remains a challenging task due to two primary reasons~\cite{ewen2022these, xu2025gaussianproperty}. First, visually similar but physically distinct regions (e.g., mud vs. asphalt, or grass vs. rock) are often difficult to distinguish, leading to brittle planning and control in complex scenes~\cite{ewen2022these, ewen2024you}. While conventional 3D mapping approaches, such as occupancy grids~\cite{elfes2002using}, signed distance fields~\cite{oleynikova2016signed}, or implicit neural representations~\cite{mildenhall2021nerf, kerbl20233d}, can provide rich geometric detail, they primarily focus on recovering object shape, and typically fail to encode physical properties or material categories.
Second, although recent work has made progress in estimating physical properties in indoor environments~\cite{zhai2024physical, xu2025gaussianproperty, park2025understanding}, outdoor scenes remain underexplored. Existing methods often focus on one or two specific physical properties of outdoor objects or terrains, such as friction~\cite{liang2022adaptiveon, ewen2022these, chen2024identifying}, pliability~\cite{weerakoon2023graspe}, or stiffness~\cite{chen2024identifying}, and are not easily generalizable to a broader set of physical attributes. 

Recent advances in 3D vision and semantic perception have increasingly focused on linking visual appearance to underlying physical attributes, enabling models to distinguish between rigid, deformable, slippery, and compliant surfaces~\cite{zhai2024physical, xu2025gaussianproperty, ewen2022these, ewen2024you, chen2024identifying}. In parallel, vision-language models (VLMs) have demonstrated the ability to capture latent physical properties, such as friction and elasticity, with multimodal inputs~\cite{zhai2024physical}. These models can qualitatively reason about forces, materials, and object dynamics from language and imagery~\cite{wang2023newton, zheng2025large, park2025understanding, xie2024deligrasp, shuai2025pugs}, highlighting their potential as semantic priors for physical inference.

However, a core challenge in estimating physical properties from visual sensors 
is managing the uncertainty inherent in both sensing and inference. Visual and depth observations are often degraded by sensor noise, lighting variation, occlusion, and calibration drift~\cite{thrun2000probabilistic, thrun2002probabilistic, tremblay2018training, chopra2024agrinerf}. This type of uncertainty, known as \textit{aleatoric uncertainty}, captures measurement noise and perceptual ambiguity~\cite{kendall2017uncertainties, cai2024evora}. Simultaneously, learning-based models trained on limited or domain-specific datasets frequently struggle to generalize to novel textures, materials, and environmental conditions~\cite{kendall2017uncertainties, kendall2018multi, hendrycks2019benchmarking}. This is referred to as \textit{epistemic uncertainty}, which reflects the model’s incomplete or imperfect knowledge of the world, often due to insufficient or biased training data~\cite{kendall2017uncertainties, cai2024evora}.

{\noindent\bf Main Contributions:}
We propose a 3D physical property estimation framework that integrates Bayesian inference into the Gaussian Splatting optimization process, treating each Gaussian primitive as a probabilistic entity whose properties are updated via posterior refinement. This enables PhysGS to estimate both point-level physical properties (e.g., friction, hardness, stiffness, density) and object-level quantities (e.g., total mass), while producing calibrated aleatoric and epistemic uncertainty.  
Our novel contributions include:

\begin{enumerate}
\item \textbf{Bayesian-Inferred Gaussian Splatting.} We embed Bayesian updates within the Gaussian Splatting pipeline, allowing each Gaussian’s physical property values to be updated through confidence-weighted posterior refinement from observations.

\item \textbf{Unified multi-property estimation across scales.} A single Bayesian formulation supports diverse physical properties, including friction, hardness, stiffness, density, and mass, at both the point level and the object level, enabling fine-grained property mapping and global aggregation within the same framework.

\item \textbf{Generality across environments and object types.} PhysGS is broadly applicable to a wide range of indoor and outdoor scenes and operates on both rigid and deformable objects, including vegetation, soil, and everyday household materials, enabling consistent physical property estimation across heterogeneous real-world settings.

\end{enumerate}

Across all datasets, including ABO-500, and a real-world friction–hardness dataset, \textit{PhysGS} achieves strong gains over prior methods such as NeRF2Physics, CLIP-based recognition, and direct VLM regression. We observe improvements of up to 61.2\% in Shore hardness error, 18.1\% in kinetic friction error, and 22.8\% in mass-density error.





\section{Related Work}

\subsection{Visual Property Fields}

A major line of work estimates physical properties by associating scene objects with open-vocabulary physical semantics, querying where specific physical property appear in observed spaces. LERF grounds CLIP embeddings~\cite{radford2021learning} within NeRF, distilling multi-scale language features into a dense, queryable 3D field that produces 3D relevancy maps for text prompts~\cite{kerr2023lerf}. Closely related open-vocabulary 3D mapping approaches propagate physical–semantic features into 3D reconstructions for zero-shot recognition and retrieval~\cite{peng2023openscene, jatavallabhula2023conceptfusion}.

With the rise of 3D Gaussian Splatting (3DGS)\cite{kerbl20233d}, several methods directly inject language or semantic features into Gaussian primitives, yielding fast, explicit, and queryable 3D fields. 
LangSplat~\cite{qin2024langsplat} distills 2D CLIP features into a 3D language field over Gaussians for open-vocabulary search.
Related efforts~\cite{guo2024semantic, hu2024semantic} assign semantic Gaussians for open-vocabulary 3D understanding, showing that explicit Gaussian fields are well-suited for encoding and rendering high-dimensional properties beyond color.

Based on visual–semantic cues, recent work has further incorporated language-context features to enhance physical property estimation~\cite{zhai2024physical, xu2025gaussianproperty}. NeRF2Physics\cite{zhai2024physical} constructs a language-embedded 3D feature space and performs zero-shot kernel regression to estimate per-point physical properties, while GaussianProperty\cite{xu2025gaussianproperty} extends this idea to 3D Gaussians. 

\subsection{Uncertainty-Aware and Probabilistic Scene Understanding}

In vision and robotics, uncertainty is typically decomposed into aleatoric (data or sensor noise) and epistemic (model or knowledge) components. This formulation has become standard for loss design, model calibration, and risk-sensitive decision making in perception systems~\cite{kendall2017uncertainties, cai2024evora, cai2025pietra}. Prior work formalizes these notions for vision-based tasks and demonstrates how to jointly learn uncertainty with outputs such as depth or segmentation, or to approximate Bayesian inference via dropout or ensembles, thereby improving robustness and out-of-distribution behavior~\cite{kendall2018multi, gal2016dropout, lakshminarayanan2017simple}.

Estimating the physical properties of real-world environments from sensors inherently involves uncertainty. Several approaches introduce probabilistic maps, tail-risk measures, and confidence-aware policies to quantify and mitigate this uncertainty~\cite{chen2024identifying, fan2021step, frey2024roadrunner, patel2024roadrunner, erni2023mem}. STEP~\cite{fan2021step} models traversability as a stochastic variable and plans using a CVaR-based risk formulation, validated across diverse field environments and the DARPA SubT Challenge. Evidential and Bayesian formulations extend this concept by outputting full distributions rather than point estimates, enabling online belief updates from new observations~\cite{cai2024evora, cai2025pietra, ewen2022these, ewen2024you}. EVORA~\cite{cai2024evora} learns evidential traction distributions, explicitly separating aleatoric and epistemic components to assess motion risk, while Ewen et al.~\cite{ewen2022these, ewen2024you} maintain joint beliefs over semantics and continuous properties to predict physical property maps (e.g. friction).

Beyond 2D perception, uncertainty has been integrated into NeRF and 3D Gaussian Splatting (3DGS) frameworks to quantify ambiguity arising from sparse views, occlusions, and under-constrained geometry~\cite{lee2025bayesian, savant2024modeling}. Recent works estimate spatial uncertainty post hoc for pre-trained NeRFs~\cite{goli2024bayes, lee2025bayesian}, propose probabilistic NeRFs~\cite{hoffman2023probnerf}, or directly model uncertainty in 3DGS~\cite{kim2025e2} via variational or evidential objectives, including dynamic and 4D settings~\cite{kim20244d}. Complementary efforts have explored uncertainty-aware on variational Gaussian splatting and SLAM pipelines that propagate uncertainty in pose and structure, demonstrating that per-Gaussian uncertainty can substantially enhance mapping robustness and downstream reasoning~\cite{savant2024modeling, hu2024cg, sandstrom2025splat}.

\section{Proposed Approach}
\label{sec:approach}

In this section, we outline our Bayesian framework for dense physical property estimation. We begin with a Dirichlet--Categorical model for fusing confidence-weighted material labels across views (Sec.~\ref{sec:preliminaries},~\ref{sec:bayesian_inference}), then extend it to continuous properties using a Normal--Inverse--Gamma prior to obtain calibrated aleatoric and epistemic uncertainty (Sec.~\ref{sec:uncertainty_fields}). We then describe how these Bayesian updates integrate with 3D Gaussian Splatting, segmentation, and VLM prompting to produce per-point property fields and object-level estimates (Sec.~\ref{sec:property_estimation}).

\begin{figure*}
    \centering
\includegraphics[width=0.85\textwidth]{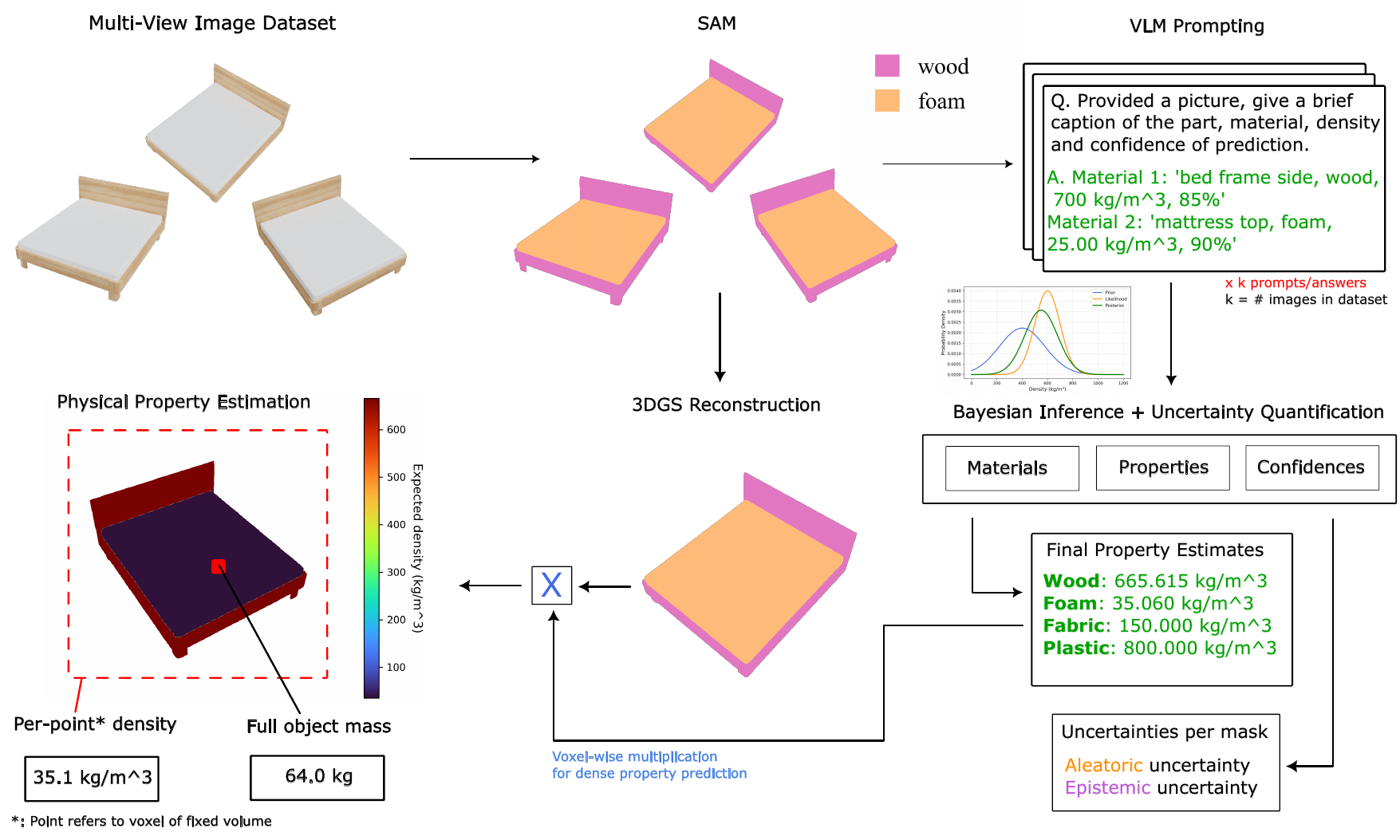}
\caption{\textit{PhysGS} architecture.
Given multi-view images, SAM provides part-level segmentations that are used for 3D Gaussian Splatting (3DGS) reconstruction. For each segmented part, a VLM produces material labels, density estimates, and confidence scores across multiple views. These observations are fused using Bayesian inference with uncertainty quantification to obtain final per-material property distributions. By propagating the estimated densities over the reconstructed 3D Gaussian field, \textit{PhysGS} predicts per-point density and full-object mass.}
\label{fig:architecture}
\vspace{-.5cm}
\end{figure*}

\subsection{Preliminaries}
\label{sec:preliminaries}

\paragraph{Dirichlet--Categorical formulation.} 
We model discrete material labels produced by the VLM using a Categorical distribution and 
place a Dirichlet prior over its parameters. The Dirichlet distribution is the conjugate prior 
to the Categorical likelihood, enabling closed-form Bayesian updates as new observations are 
incorporated across views.

The Categorical distribution parameterized by $\boldsymbol{\theta} \in [0,1]^K$ represents the probability that an observation belongs to class $i$:
\begin{equation}
    f(z=i \mid \boldsymbol{\theta}) = \theta_i.
    \label{eq:categorical}
\end{equation}
The Dirichlet distribution defines a continuous $K$-variate prior over $\boldsymbol{\theta}$, parameterized by $\boldsymbol{\alpha} \in \mathbb{R}_{>0}^K$, as
\begin{equation}
    f(\boldsymbol{\theta} \mid \boldsymbol{\alpha}) =
    \frac{\Gamma(\sum_{k=1}^K \alpha_k)}
         {\prod_{k=1}^K \Gamma(\alpha_k)}
    \prod_{k=1}^K \theta_k^{\alpha_k - 1},
    \label{eq:dirichlet_pdf}
\end{equation}
where $\Gamma(\cdot)$ is the Gamma function.  

Given a set of $n$ observed material labels 
$\mathcal{Z} = \{ z_1, \dots, z_n \}$ drawn from a Categorical distribution, 
the posterior predictive probability that a new observation belongs to material class $i$ is
\begin{equation}
    f(z=i \mid \mathcal{Z}, \boldsymbol{\alpha}) =
    \int_{\boldsymbol{\theta}}
        f(z=i \mid \boldsymbol{\theta}) \,
        f(\boldsymbol{\theta} \mid \mathcal{Z}, \boldsymbol{\alpha})
    \, d\boldsymbol{\theta}.
    \label{eq:posterior_integral}
\end{equation}
Using conjugacy of the Dirichlet prior and Categorical likelihood, 
the integral simplifies to the closed-form expression
\begin{equation}
    f(z=i \mid Z, \boldsymbol{\alpha}) = 
    \frac{\tilde{\alpha}_i}{\sum_{j=1}^{K} \tilde{\alpha}_j},
    \label{eq:posterior_class}
\end{equation}
where the posterior parameters are recursively updated as
\begin{equation}
    \tilde{\alpha}_i 
    \leftarrow 
    \alpha_i(0) + 
    \sum_{m:\, c_m = i} \lambda\, p_m,
    \label{eq:dirichlet_update}
\end{equation}
with $\lambda$ controlling the evidence strength contributed by each observation and $p_m$ 
denoting the confidence provided by the VLM for the $m$-th prediction.

\subsection{Bayesian Inference for Material Property Estimation}
\label{sec:bayesian_inference}
We introduce our hierarchical Bayesian framework for estimating material-specific physical properties from confidence-weighted observations. Building on the Dirichlet--Categorical model of~\cite{ewen2024you}, we extend it with a continuous posterior to jointly infer material class and properties such as friction, density, and hardness.

\vspace{3pt}
\noindent\textbf{Continuous property estimation.}
While the Dirichlet--Categorical formulation governs the discrete class probabilities, 
we also require an estimate of the continuous physical property $\psi$ associated with each material. 
For each material class $i$, we maintain confidence-weighted accumulators that enable 
incremental computation of the first and second moments using a running mean and variance formulation 
proposed by ~\cite{west1979updating} and generalized by ~\cite{pebay2008formulas}:
\begin{equation}
    W_i = \sum_m p_m, \qquad
    S_i = \sum_m p_m\, \psi_m, \qquad
    Q_i = \sum_m p_m\, \psi_m^2,
    \label{eq:weighted_moments}
\end{equation}
representing the total weight, first moment, and second moment, respectively. 
These accumulators allow efficient online updates without requiring access to past observations,
which is particularly beneficial in streaming or on-the-fly reconstruction settings. 

The posterior mean and variance for material $i$ are then estimated as
\begin{equation}
    \mu_i = \frac{S_i}{W_i}, 
    \qquad
    \sigma_i^2 = 
    \max\!\left(
        \frac{Q_i}{W_i} - \mu_i^2, \, \epsilon
    \right),
    \label{eq:posterior_mean_var}
\end{equation}
yielding a Gaussian posterior
\begin{equation}
    p(\psi_i \mid Z) = 
    \mathcal{N}(\mu_i, \sigma_i^2),
    \label{eq:gaussian_posterior}
\end{equation}
which represents the system’s belief over the continuous physical property for material $i$ given all confidence-weighted evidence $Z$. This formulation integrates naturally with the Dirichlet update by providing confidence-weighted, incremental, and uncertainty-aware refinement as new observations become available.

\vspace{3pt}
\noindent\textbf{Hierarchical posterior.}
The resulting model is hierarchical in nature, jointly estimating the discrete material identity $z$ and continuous physical property $\psi$:
\begin{equation}
    p(z, \psi \mid Z, \boldsymbol{\alpha}) 
    = p(\psi \mid z, Z)\, p(z \mid Z, \boldsymbol{\alpha}),
    \label{eq:hierarchical_posterior}
\end{equation}
where $p(z \mid Z, \boldsymbol{\alpha})$ is the Dirichlet–Categorical posterior 
and $p(\psi \mid z, Z)$ is the Gaussian posterior.  
Applying the Law of Total Probability as in~\cite{ewen2024you}, 
the overall predictive distribution over physical properties is
\begin{equation}
    f(\psi \mid Z, \boldsymbol{\alpha}) =
    \sum_{i=1}^{K}
    f(\psi \mid z=i)\, f(z=i \mid Z, \boldsymbol{\alpha}).
    \label{eq:total_probability}
\end{equation}

Substituting Eq.~\eqref{eq:gaussian_posterior} into Eq.~\eqref{eq:total_probability}
gives a closed-form multimodal Gaussian mixture for the predicted material properties:
\begin{equation}
    f(\psi \mid Z, \boldsymbol{\alpha}) =
    \sum_{i=1}^{K}
    \frac{\tilde{\alpha}_i}{\sum_{j=1}^{K} \tilde{\alpha}_j}\,
    \mathcal{N}(\mu_i, \sigma_i^2).
    \label{eq:mixture_gaussian}
\end{equation}
This mixture formulation expresses the full posterior as a 
weighted sum of unimodal Gaussian components, where each mode corresponds to a material class 
and is weighted by its recursively updated class likelihood from the Dirichlet posterior.


\subsection{Uncertainty-Aware Property Fields}
\label{sec:uncertainty_fields}

\vspace{3pt}
\noindent\textbf{Uncertainty modeling via the Normal–Inverse–Gamma prior.}
For each material $i$, the joint prior over the mean $\mu_i$
and variance $\sigma_i^2$ of the property $\psi$ is given by
\begin{equation}
\begin{aligned}
p(\mu_i, \sigma_i^2 \mid \tau_i, \kappa_i, \alpha_i, \beta_i)
\;=\;&
\mathcal{N}\!\left(
    \mu_i \,\middle|\, \tau_i, \tfrac{\sigma_i^2}{\kappa_i}
\right)
\\
&\mathrm{Inv\text{-}Gamma}\!\left(
    \sigma_i^2 \,\middle|\, \alpha_i, \beta_i
\right),
\end{aligned}
\label{eq:nig_prior_material}
\end{equation}

where 
$\tau_i$ denotes the prior mean, 
$\kappa_i$ controls the precision on $\mu_i$ (i.e., the strength of accumulated evidence),
and $(\alpha_i, \beta_i)$ are the shape and scale parameters governing the uncertainty in $\sigma_i^2$.

\vspace{3pt}
\noindent\textbf{Predictive Uncertainty update.}
Given a new observation $\psi_m$ associated with material class $i$
and its confidence $p_m$, the posterior parameters
$(\tilde{\tau}_i, \tilde{\kappa}_i, \tilde{\alpha}_i, \tilde{\beta}_i)$
can be updated in closed-form, allowing sequential fusion of confidence-weighted evidence without storing past data.
This conjugate formulation provides closed-form expressions for the predictive
mean and variance of the property $\psi_i$.

The total predictive uncertainty decomposes into two components:
\begin{equation}
    \mathrm{Var}[\psi_i]
    =
    \underbrace{\mathbb{E}[\sigma_i^2]}_{\text{aleatoric}}
    +
    \underbrace{\mathrm{Var}[\mu_i]}_{\text{epistemic}}.
    \label{eq:uncertainty_decomposition}
\end{equation}
The first term represents \textit{aleatoric uncertainty}, which arises from
inherent noise in the observations and variability within each material class.
The second term represents \textit{epistemic uncertainty}, corresponding to
uncertainty in the estimated mean that decreases as more high-confidence
evidence is incorporated.
For implementation, we compute these moments directly from the NIG parameters:
\begin{equation}
    \mathbb{E}[\sigma_i^2] = 
    \frac{\tilde{\beta}_i}{\tilde{\alpha}_i - 1},
    \qquad
    \mathrm{Var}[\mu_i] =
    \frac{\mathbb{E}[\sigma_i^2]}{\tilde{\kappa}_i}.
    \label{eq:uncertainty_moments}
\end{equation}

The resulting aleatoric, epistemic, and total predictive uncertainties provide
interpretable measures of confidence in both the per-class property estimates
and the overall reconstruction.  Regions with high aleatoric uncertainty reflect
sensor or perceptual noise, while high epistemic uncertainty indicates
insufficient or conflicting evidence about the underlying material properties.

\subsection{Learning Semantics and Physical Properties}
\label{sec:property_estimation}

\vspace{3pt}
\noindent\textbf{Semantic Segmentation.}
Given a multi-view image dataset, we employ SAM~\cite{kirillov2023segment} to produce pixel-accurate masks that decompose each object into hierarchical levels (whole, part, and sub-part), facilitating fine-grained semantic understanding. 
The model outputs multiple candidate masks at different granularities, which we refine by discarding redundant or low-confidence predictions using SAM’s built-in IoU and stability measures. 
The resulting segmentation maps capture precise object boundaries and semantically coherent regions, forming the basis for our downstream physical property estimation.

\vspace{3pt}
\noindent\textbf{VLM Prompting.}
For each segmented image, we construct a vision--language prompt comprising a triplet of images arranged side by side, following the design of~\cite{xu2025gaussianproperty}. The left image presents the complete object, the middle image overlays the segmentation mask, and the right image isolates the masked region of interest. Given an input image $I$, this process yields $k$ visual prompts corresponding to the $k$ masks predicted by SAM.  
We additionally condition the VLM with a structured textual query that instructs it to (i) provide a concise caption of the segmented part, (ii) identify its predominant material, and (iii) infer relevant physical properties such as friction, density, etc. The model is further asked to report a normalized confidence score within $[0,1]$, representing its belief in the prediction.

\vspace{3pt}
\noindent\textbf{3D Gaussian Splatting.}
Given the VLM responses and the refined physical property estimates from our Bayesian inference scheme, we construct a material legend assigning each material a unique color. The corresponding scene images are recolored accordingly and used as semantic inputs for 3DGS reconstruction. This yields a semantic splat that supports dense property inference, such as mass estimation.

\vspace{3pt}
\noindent\textbf{Physical Property Estimation.}
Using the reconstructed 3D Gaussian field and inferred material properties, we perform per-point and dense physical property estimation. Each voxel is associated with a predicted property value (e.g., friction or density), enabling spatial queries for per-point properties or integration over the volume to obtain aggregate measures such as total mass.


\section{Experiments and Results}


\begin{figure*}
    \centering
\includegraphics[width=0.85\textwidth]{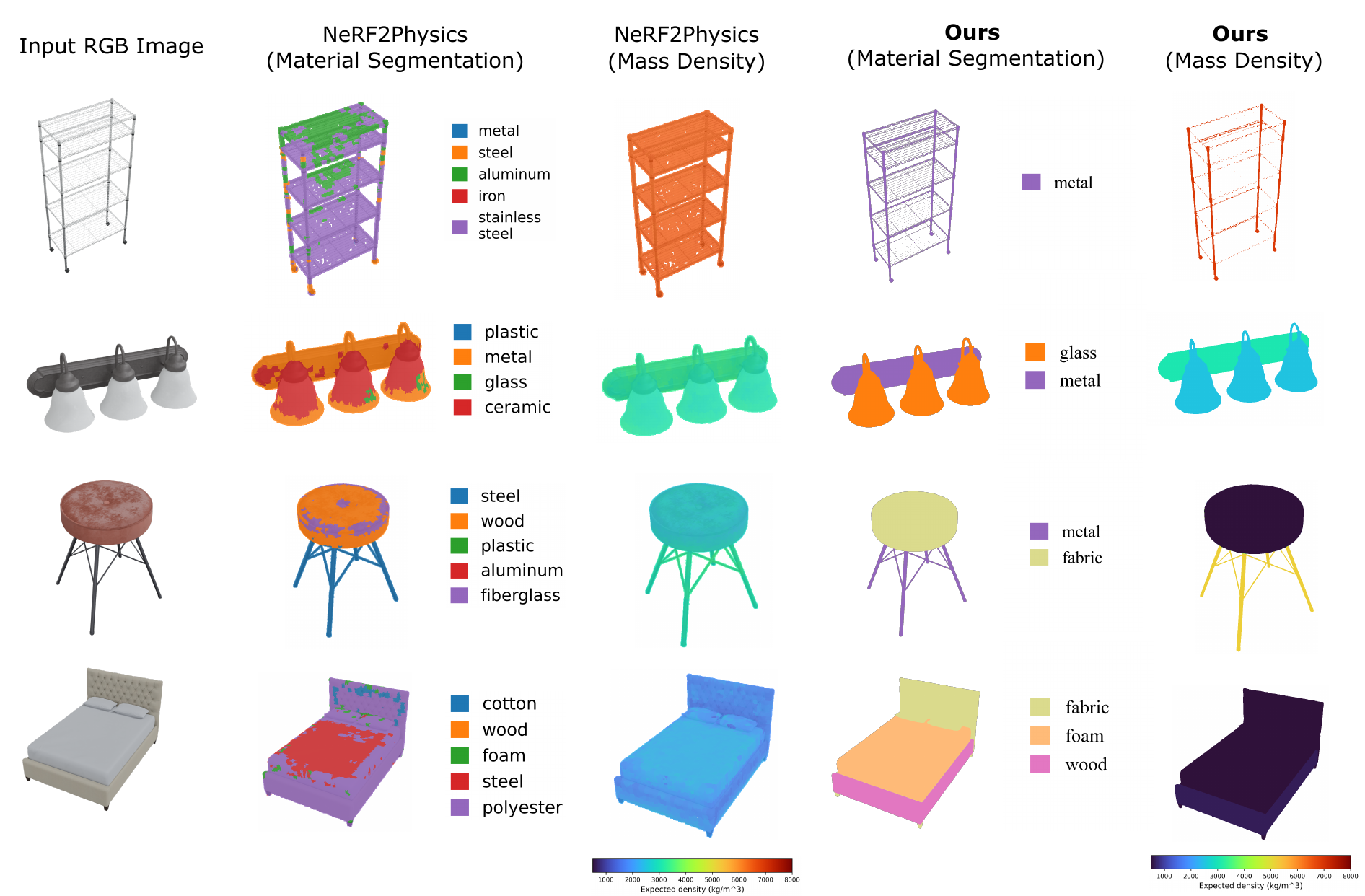}
\caption{Qualitative comparison on the ABO-500 dataset. For each object, we show the input RGB image, material segmentation and mass-density predictions from NeRF2Physics, and the corresponding results from our method. \textit{PhysGS} produces cleaner material segmentation with fewer artifacts compared to NeRF2Physics and more consistent part boundaries, and yields sharper, more plausible mass-density 
fields across diverse object categories.}
\vspace{-.5cm}
\label{fig:abo_qualitative}
\end{figure*}

\subsection{Implementation Details}

We employ the \texttt{splatfacto-big} variant of Nerfstudio~\cite{tancik2023nerfstudio} for 3D Gaussian Splatting, using default parameters except for a random scale of $2.0$ and a random background color. Each scene is trained for $20{,}000$ iterations on an NVIDIA RTX~A5000 GPU.  

For image segmentation, we use SAM~\cite{kirillov2023segment} to obtain whole, part, and sub-part material decompositions. Material property estimation is performed using GPT-5 as the vision–language model (VLM), conditioned on structured visual–text prompts derived from the segmented images.

\subsection{Mass Estimation}

\vspace{3pt}
\noindent\textbf{Dataset.}
We employ the ABO dataset~\cite{collins2022abo} for evaluating mass prediction, which includes a large set of consumer products listed on Amazon along with multi-view imagery, segmentation masks, physical measurements, and metadata. Specifically, we make use of the representative multi-view benchmark ABO-500 curated by ~\cite{zhai2024physical}, which selects a balanced subset of 500 items from the entire ABO dataset. It is divided into 300 training, 100 validation, and 100 testing instances.

\vspace{3pt}
\noindent\textbf{Metrics.}
Following prior work on visual mass estimation~\cite{standley2017image2mass}, we evaluate using four complementary metrics that measure both absolute and relative error between the predicted mass $\hat{m}$ and the ground-truth mass $m$:
\begin{itemize}
    \item Absolute Difference Error (ADE): $|m - \hat{m}|$,
    \item Absolute Log Difference Error (ALDE): $|\ln m - \ln \hat{m}|$,
    \item Absolute Percentage Error (APE): $\big|\frac{m - \hat{m}}{m}\big|$,
    \item Minimum Ratio Error (MnRE): $\min\left(\frac{m}{\hat{m}}, \frac{\hat{m}}{m}\right)$.
\end{itemize}

\vspace{3pt}
\noindent\textbf{Baselines.}
We compare our system against several visual and multimodal baselines on the ABO-500 dataset:
\begin{itemize}
    \item \textbf{Image2mass}~\cite{standley2017image2mass}: a CNN that infers mass directly from RGB images and 3D bounding box dimensions.
    \item \textbf{2D CNN}: a lightweight regression model built upon a frozen ResNet50~\cite{he2016deep} backbone, fine-tuned with additional layers for scalar mass prediction.
    \item \textbf{LLaVA}~\cite{liu2023visual}: a vision-language model designed for instruction following.
    \item \textbf{NeRF2Physics}~\cite{zhai2024physical}: a NeRF-based approach that jointly estimates 3D geometry and per-point physical properties such as density, friction, and stiffness. It predicts mass by integrating predicted density across the reconstructed volume.
\end{itemize}

\vspace{3pt}
\noindent\textbf{Qualitative Results.}
Figure~\ref{fig:abo_qualitative} presents qualitative results on the ABO-500 dataset. For each object, we compare the material segmentation and mass-density predictions from NeRF2Physics with those produced by \textit{PhysGS}. Our method yields substantially cleaner material segmentations with fewer spurious labels and more coherent part boundaries, while also producing sharper and more stable mass-density fields. These improvements are consistent across a wide range of object categories, demonstrating the advantage of combining vision-language priors with Bayesian inference over 3D Gaussian splats.

\vspace{3pt}
\noindent\textbf{Quantitative Results.}
We evaluate the accuracy of our method on mass estimation using the ABO-500 test set (100 objects). As shown in Table~\ref{tab:abo_table}, traditional 2D methods such as Image2mass ~\cite{standley2017image2mass} and 2D CNNs exhibit high mass estimation error due to their inability to capture 3D structure or material composition. VLM approaches (e.g. LLaVA~\cite{liu2023visual}) show similar limitations, producing  noisy predictions that vary across views. NeRF2Physics improves accuracy by exploiting neural radiance fields, and achieves the best ALDE (0.771) and MnRE (0.552) among existing baselines. \textit{PhysGS} achieves the best performance on two key metrics: it reduces ADE from 8.730 to 8.254, corresponding to a 5.5\% improvement, and reduces APE from 1.061 to 0.819, a substantial 22.8\% improvement over NeRF2Physics.

\vspace{3pt}
\noindent\textbf{Ablation Study.}
Table~\ref{tab:abo_ablation} shows that incorporating Bayesian inference yields clear gains over both NeRF2Physics and our non-Bayesian variant. Updating material and property beliefs across additional views reduces ADE by 5.6\% and improves APE by 6.4\% compared to the version without Bayesian updates. These improvements demonstrate that aggregating multi-view evidence to refine the posterior distribution leads to more accurate mass estimation, confirming the benefit of treating physical properties as latent variables that are iteratively updated rather than fixed from single-view predictions.

\begin{table}[t]
\centering
\setlength{\tabcolsep}{1.8pt} 
\caption{Mass estimation on ABO-500 test set (100 objects). ADE is measured in kilograms. \textbf{Bold}: best model.}
\label{tab:abo_table}
\begin{tabular}{lcccc}
\toprule
\textbf{Method} & \textbf{ADE} ($\downarrow$) & \textbf{ALDE} ($\downarrow$) & \textbf{APE} ($\downarrow$) & \textbf{MnRE} ($\uparrow$) \\
\midrule
Image2mass~\cite{standley2017image2mass} & 12.496 & 1.792 & 0.976 & 0.341 \\
2D CNN & 15.431 & 1.609 & 14.459 & 0.362 \\
LLaVA~\cite{liu2023visual} & 17.328 & 1.893 & 1.837 & 0.306 \\
NeRF2Physics~\cite{zhai2024physical} & 8.730 & \textbf{0.771} & 1.061 & \textbf{0.552} \\
\midrule
\textbf{Ours} & \textbf{8.254} & 0.999 & \textbf{0.819} & 0.474 \\
\bottomrule
\end{tabular}
\vspace{-0.3cm}
\end{table}

\begin{table}[t] \centering 
\setlength{\tabcolsep}{1.8pt} 
\caption{Ablation study for mass estimation on ABO-500 val set (100 objects). ADE is measured in kilograms. BI refers to Bayesian Inference.
\label{tab:abo_ablation}
\textbf{Bold}: best model.} \begin{tabular}{lcccc} \toprule \textbf{Method} & \textbf{ADE} ($\downarrow$) & \textbf{ALDE} ($\downarrow$) & \textbf{APE} ($\downarrow$) & \textbf{MnRE} ($\uparrow$) \\ \midrule NeRF2Physics~\cite{zhai2024physical} & 9.786 & \textbf{0.61} & 0.931 & \textbf{0.609} 
\\ Ours (w/o BI) & 9.728 & 0.770 & 0.717 & 0.561 \\ 
\midrule
\textbf{Ours (with BI)} & \textbf{9.187} & 0.827 & \textbf{0.715} & 0.539 \\ \bottomrule 
\end{tabular}
\vspace{-.5cm}
\end{table}

\begin{figure}[t!]
    \centering
\includegraphics[width=0.4\textwidth]{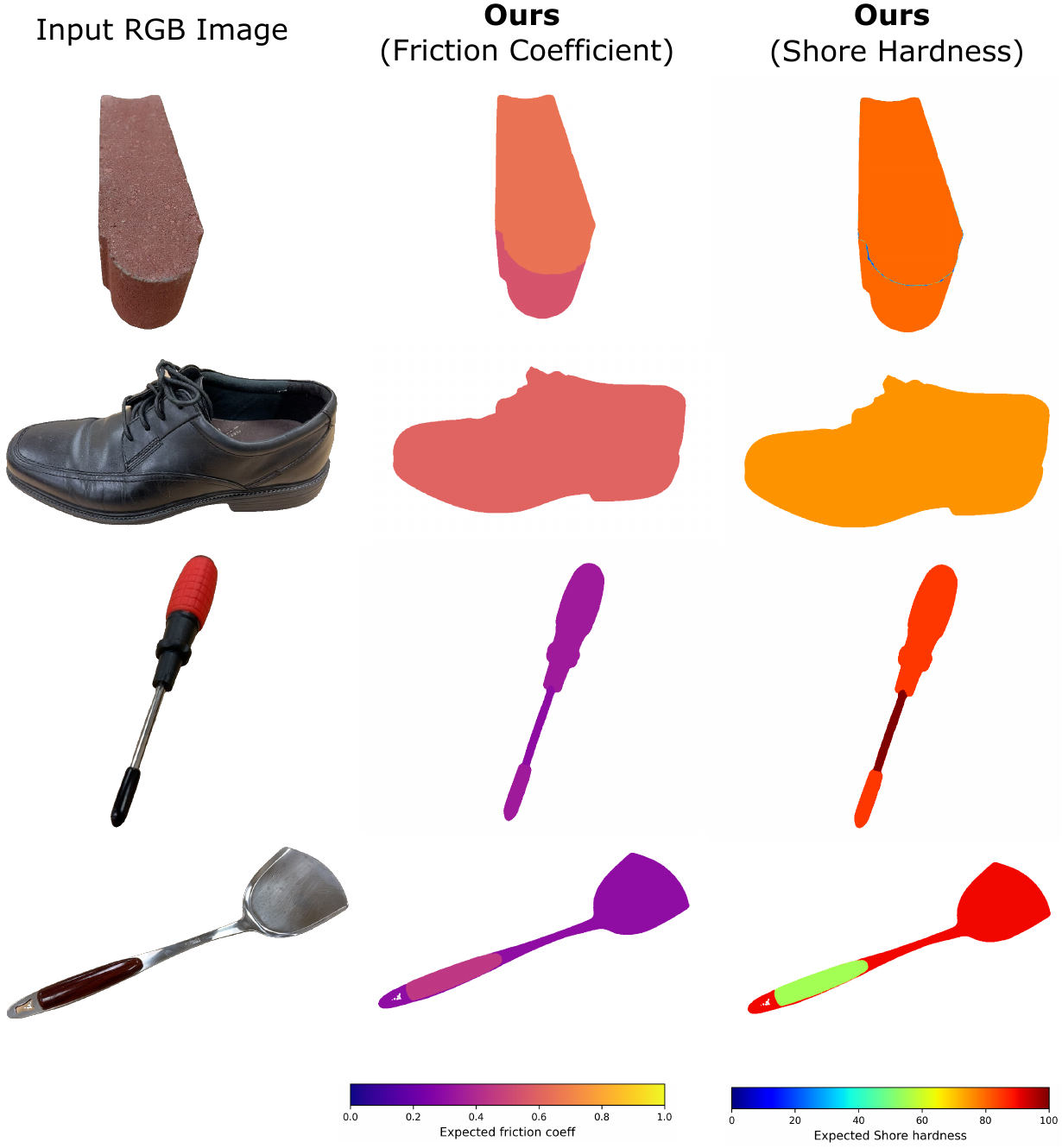}
\caption{Qualitative results for the friction and hardness dataset. Given a single RGB view, \textit{PhysGS} predicts dense friction coefficients and Shore hardness values for a variety of household 
objects. The resulting property fields are spatially smooth, physically plausible, and consistent across diverse materials and geometries.}
\vspace{-.5cm}
\label{fig:friction_hardness}
\end{figure}

\begin{table*}[t]
\centering
\setlength{\tabcolsep}{2.4pt}
\caption{Estimation of per-point Shore hardness (left) and kinetic friction coefficient (right) on the real-world dataset. \textbf{Bold}: best model.}
\label{tab:hardness_friction_results}
\begin{tabular}{lccccccccc}
\toprule
\multicolumn{5}{c}{\textbf{Shore Hardness (31 points, 11 objects)}} & \multicolumn{5}{c}{\textbf{Kinetic Friction (6 points, 6 objects)}} \\
\cmidrule(lr){1-5} \cmidrule(lr){6-10}
\textbf{Method} & \textbf{ADE}~($\downarrow$) & \textbf{ALDE}~($\downarrow$) & \textbf{APE}~($\downarrow$) & \textbf{MnRE}~($\uparrow$) &
\textbf{Method} & \textbf{ADE}~($\downarrow$) & \textbf{ALDE}~($\downarrow$) & \textbf{APE}~($\downarrow$) & \textbf{MnRE}~($\uparrow$) \\
\midrule
GPT-4V~\cite{yang2023dawn} & 32.752 & 0.330 & 0.304 & 0.758 & 
GPT-4V~\cite{yang2023dawn} & 0.209 & 0.430 & 0.549 & 0.692 \\
CLIP~\cite{radford2021learning} & 32.857 & 0.294 & 0.266 & 0.774 &
CLIP~\cite{radford2021learning} & 0.222 & 0.455 & 0.602 & 0.654 \\
NeRF2Physics~\cite{zhai2024physical} & 34.295 & 0.315 & 0.276 & 0.765 &
NeRF2Physics~\cite{zhai2024physical} & 0.155 & 0.321 & \textbf{0.360} & 0.736 \\
\midrule
\textbf{Ours} & \textbf{12.721} & \textbf{0.193} & \textbf{0.222} & \textbf{0.839} & 
\textbf{Ours} & \textbf{0.131} & \textbf{0.263} & 0.365 & \textbf{0.805} \\
\bottomrule
\end{tabular}
\label{tab:combined_hardness_friction}
\vspace{-.3cm}
\end{table*}

\begin{figure*}
    \centering
\includegraphics[width=0.9\textwidth]{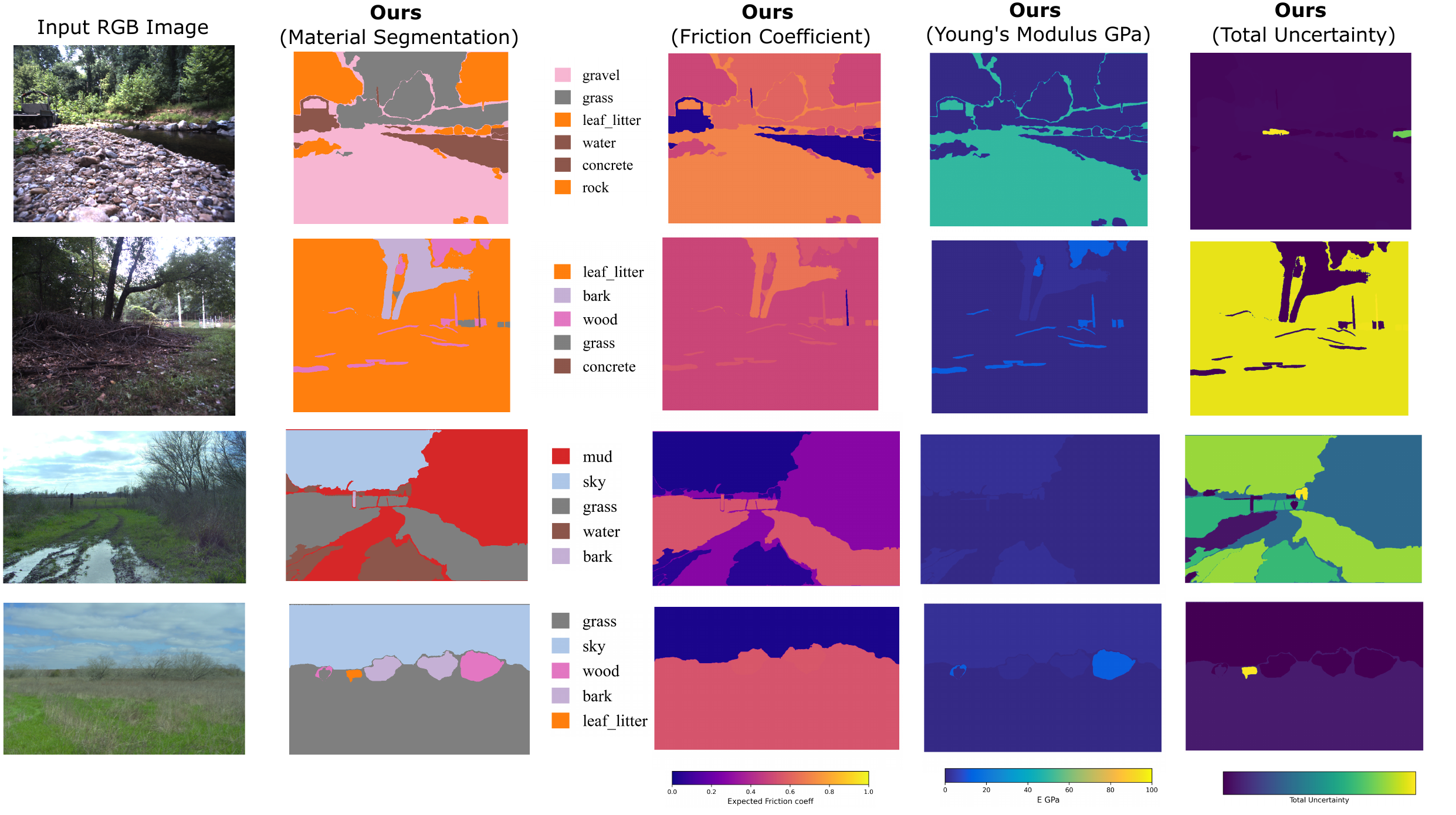}
\caption{Outdoor scene results on real environments. From a single RGB view, \textit{PhysGS} predicts material segmentation, friction coefficients, Young’s modulus, and total uncertainty (\textit{aleatoric} + \textit{epistemic}). The method captures broad material variations across natural terrain and vegetation while producing pixel-wise physical property estimates with associated confidence.  Higher total uncertainty in Rows~2 and~3 corresponds to scenes with dense clutter and visually ambiguous regions, where SAM provides less precise part-level masks (e.g., separating leaf litter from wood or mud from grass). Rows~1 and~4 exhibit lower uncertainty due to clearer material boundaries and more uniform regions.}
\label{fig:outdoor_scene}
\vspace{-.5cm}
\end{figure*}

\subsection{Friction and Hardness Estimation}

\vspace{3pt}
\noindent\textbf{Dataset.}
To evaluate our model’s ability to infer dense, spatially varying physical properties within objects, we leverage the friction and hardness dataset containing paired image and real-world measurement data, curated by ~\cite{zhai2024physical}.
This dataset includes 15 household objects captured across 13 scenes, using multi-view RGB images paired with per-point measurements of kinetic friction coefficient and Shore A/D hardness. 

\vspace{3pt}
\noindent\textbf{Metrics.}
We report the same evaluation metrics for per-point friction and hardness estimation used in evaluation for mass estimation as above: ADE, ALDE, APE, MnRE.

\vspace{3pt}
\noindent\textbf{Baselines.}
As before, we compare our method to several visual and multimodal baselines.
\begin{itemize}
\item \textbf{GPT-4V} \cite{yang2023dawn}: A large vision–language model capable of processing masked regions in its prompt.
\item \textbf{CLIP} \cite{radford2021learning}: A vision–language baseline that uses global CLIP embeddings from the canonical view of the scene, rather than the fused multi-view patch features used in our method. This baseline evaluates how well static visual–semantic representations can generalize to physical property prediction.
\item \textbf{NeRF2Physics} \cite{zhai2024physical}: Same as in mass estimation baseline.
\end{itemize}

\vspace{3pt}
\noindent\textbf{Qualitative Results.}
Figure~\ref{fig:friction_hardness} presents qualitative results for friction and Shore hardness estimation on real objects. From a single RGB view, \textit{PhysGS} produces dense per-point friction and hardness fields across materials such as rubber, leather, plastic, and metal. The predictions capture fine-grained material differences and exhibit clean boundaries between regions with distinct friction and hardness characteristics, reflecting the system’s ability to localize subtle variations in contact and deformation properties.

\vspace{3pt}
\noindent\textbf{Quantitative Results.}
Table~\ref{tab:hardness_friction_results} reports results for per-point Shore A/D hardness and kinetic friction estimation on our real-world dataset. Across all hardness metrics, PhysGS achieves substantial gains over existing approaches. Relative to the best baseline, our method reduces ADE by 61.2\%, lowers ALDE by 34.4\%, and decreases APE by 16.5\%, while improving MnRE by 8.4\%.  For kinetic friction, PhysGS again outperforms the strongest baseline (NeRF2Physics) on the majority of metrics, reducing ADE by 15.5\% and ALDE by 18.1\%, and increasing MnRE by 9.4\%. These gains highlight the benefit of integrating multi-view evidence through Bayesian inference, which refines the posterior distribution of material properties beyond what single-view or deterministic models can infer.

\subsection{Applications: Outdoor Scene Understanding}
\textit{\ours{}} can also estimate physical properties of outdoor environments, such as friction and stiffness, which are important for reasoning about natural, vegetation-rich terrain~\cite{chen2024identifying}. Our method also provides per-pixel uncertainty (aleatoric + epistemic) for these estimates. We demonstrate this capability on the RUGD~\cite{wigness2019rugd} and RELLIS-3D~\cite{jiang2021rellis} datasets (Figure~\ref{tab:hardness_friction_results}), which contain challenging outdoor scenes where accurate physical property prediction and uncertainty assessment are essential for interpreting complex terrain.
\section{Conclusion, Limitations and Future Work}
\label{sec:conclusion}

We presented \textit{PhysGS}, a Bayesian-inferred 3D Gaussian Splatting framework for estimating dense physical properties from RGB images and vision-language priors. Across indoor and outdoor real-world datasets, \textit{PhysGS} achieves consistent gains over existing approaches. On ABO-500, our method improves mass estimation by 5.5\% in ADE and 22.8\% in APE. \textit{PhysGS} also reduces Shore hardness error by up to 61.2\% and kinetic friction error by up to 18.1\% relative to the strongest baselines. Outdoor experiments on RUGD and RELLIS-3D further show that the method generalizes to complex natural environments, capturing material segmentation, friction, stiffness, and uncertainty.

A primary limitation of \textit{PhysGS} lies in its sensitivity to segmentation quality. When part-level masks merge visually similar materials or fail to isolate fine-grained regions, the downstream physical property estimates inherit this ambiguity, reducing material separation and increasing predictive uncertainty. This effect is visible in Fig.~\ref{fig:outdoor_scene}, where cluttered outdoor regions lead to less precise SAM masks and correspondingly higher total uncertainty. Future work may incorporate VLM-guided segmentation refinement or confidence-based mask filtering to automatically reject low-quality masks and preserve fine-grained material structure.

{
    \small
    \bibliographystyle{ieeenat_fullname}
    \bibliography{main}
}

\clearpage
\setcounter{page}{1}
\maketitlesupplementary

\section*{A. Full Bayesian and Uncertainty Formulation}
\addcontentsline{toc}{section}{A. Full Bayesian and Uncertainty Formulation}

\subsection*{A.1. Observation Model}

For completeness, we describe the  observation model used in \textit{PhysGS}. Each observation corresponds to a segmented region of the scene and contains semantic (material) and physical (property) information extracted from the vision--language model (VLM). The role of these observations in the Bayesian updates is described in Sec.~\ref{sec:approach}.

\vspace{3pt}
\noindent\textbf{Observation tuple.}
For the $m$-th segmented region, we define
\begin{equation}
    \mathcal{O}_m = \big(c_m,\; p_m,\; \psi_m\big),
\end{equation}
where $c_m$ is the predicted material class, $p_m$ is the confidence produced by the VLM, and $\psi_m$ is the predicted physical property (e.g., friction, hardness, stiffness, density).  
Across multiple views, the full set of observations is
\begin{equation}
    Z = \{\mathcal{O}_1, \dots, \mathcal{O}_M\}.
\end{equation}

The tuple $(c_m, p_m, \psi_m)$ constitutes a noisy measurement of the latent variables $(z_m, \mu_{z_m}, \sigma_{z_m}^2)$. In particular, the predicted class $c_m$ serves as a noisy proxy for the true (unobserved) material label $z_m$, while the VLM estimate $\psi_m$ provides a noisy observation of the underlying material-specific physical property whose distribution is governed by $(\mu_{z_m}, \sigma_{z_m}^2)$. These observed quantities supply the confidence-weighted evidence used in the Bayesian updates that follow.

\subsection*{A.2. Dirichlet--Categorical Posterior}

The material fusion process follows the Dirichlet--Categorical formulation introduced in Sec.~\ref{sec:preliminaries}.  
The Categorical likelihood and Dirichlet prior correspond to Eqs.~(\ref{eq:categorical})--(\ref{eq:dirichlet_pdf}). 

The posterior Dirichlet parameters update as in Eq.~(\ref{eq:dirichlet_update}):
\begin{equation}
    \tilde{\alpha}_i
    = \alpha_i(0) + \sum_{m:\, c_m=i} \lambda\, p_m.
\end{equation}

The resulting posterior predictive distribution over material classes is given in Eq.~(\ref{eq:posterior_class}).

\subsection*{A.3. Continuous Property Estimation}

Continuous physical properties are fused using confidence-weighted running moments as introduced in Sec.~\ref{sec:bayesian_inference} of the main paper.  
The accumulators $W_i$, $S_i$, and $Q_i$ match Eq.~(\ref{eq:weighted_moments}) in the main paper.

The posterior mean and variance follow Eq.~(\ref{eq:posterior_mean_var}) in the main paper:
\begin{equation}
    \mu_i = \frac{S_i}{W_i},
    \qquad
    \sigma_i^2 = 
    \max\!\left(\frac{Q_i}{W_i} - \mu_i^2,\; \epsilon\right).
\end{equation}

This defines the Gaussian posterior $p(\psi_i \mid Z)$ shown in Eq.~(\ref{eq:gaussian_posterior}) of the main paper.

\subsection*{A.4. Mixture Formulation}

Marginalizing over discrete material classes using the hierarchical model in Sec.~\ref{sec:bayesian_inference} leads directly to the mixture distribution shown in Eq.~(\ref{eq:mixture_gaussian}), combining material probabilities with the class-conditional Gaussian property estimates.

\subsection*{A.5. Normal--Inverse--Gamma Posterior}

We extend our continuous estimator with the Normal--Inverse--Gamma (NIG) prior introduced in Sec.~\ref{sec:uncertainty_fields}.  
The joint prior over $(\mu_i, \sigma_i^2)$ matches Eq.~(\ref{eq:nig_prior_material}).

Given a weighted observation $(\psi_m, p_m)$, the closed-form posterior updates (Eqs.~(\ref{eq:uncertainty_decomposition})--(\ref{eq:uncertainty_moments})) are:
\begin{align}
    \tilde{\kappa}_i &= \kappa_i + p_m, \\
    \tilde{\tau}_i &= \frac{\kappa_i \tau_i + p_m \psi_m}{\kappa_i + p_m}, \\
    \tilde{\alpha}_i &= \alpha_i + \tfrac{p_m}{2}, \\
    \tilde{\beta}_i &= \beta_i
        + \frac{p_m \kappa_i (\psi_m - \tau_i)^2}{2(\kappa_i + p_m)}.
\end{align}

\subsection*{A.6. Predictive Uncertainty}

The decomposition of predictive uncertainty into aleatoric and epistemic components follows Eq.~(\ref{eq:uncertainty_decomposition}).
The predictive moments correspond directly to Eq.~(\ref{eq:uncertainty_moments}):
\begin{equation}
    \mathbb{E}[\sigma_i^2] = \frac{\tilde{\beta}_i}{\tilde{\alpha}_i - 1},
    \qquad
    \mathrm{Var}[\mu_i] = \frac{\mathbb{E}[\sigma_i^2]}{\tilde{\kappa}_i}.
\end{equation}

Aleatoric uncertainty reflects inherent variability within a material class, while epistemic uncertainty captures uncertainty in the estimated mean due to limited or inconsistent evidence.

\subsection*{A.7. MMSE Estimate}

As shown in Eq.~(\ref{eq:posterior_mean_var}), the posterior mean $\mu_i$ is the minimum mean-square-error (MMSE) estimator:
\begin{equation}
    \hat{\psi}_i = \mu_i.
\end{equation}

This corresponds to the property value that minimizes expected squared error and is therefore used as the single representative estimate for each material class.

\subsection*{A.8. Full Probabilistic Model}
\label{sec:full_prob_model}

The complete hierarchical model underlying \textit{PhysGS} is summarized in Sec.~\ref{sec:approach} of the main paper and depicted in Fig.~\ref{fig:architecture}.  
For completeness, we restate the probabilistic structure:

\begin{align}
    \boldsymbol{\theta} &\sim \mathrm{Dirichlet}(\boldsymbol{\alpha}(0)), \\
    z_m &\sim \mathrm{Categorical}(\boldsymbol{\theta}), \\
    (\mu_i, \sigma_i^2) &\sim \mathrm{NIG}(\tau_i,\kappa_i,\alpha_i,\beta_i), \\
    \psi_m &\sim \mathcal{N}(\mu_{z_m}, \sigma_{z_m}^2).
\end{align}

This formulation provides the full probabilistic backbone through which \textit{PhysGS} jointly infers materials, continuous properties, and calibrated uncertainty. First, a Dirichlet prior is placed over the material probabilities~$\boldsymbol{\theta}$, reflecting initial uncertainty about the frequency of each material class. Each segmented region then draws a material label~$z_m$ from this Categorical distribution. For every material class~$i$, the mean and variance of its physical property are modeled using a Normal--Inverse--Gamma (NIG) prior, which captures both uncertainty in the material’s typical property value and its intrinsic variability. Finally, the observed physical property~$\psi_m$ for region~$m$ is sampled from the Gaussian distribution associated with its material label. Together, this hierarchy defines how materials and their continuous properties jointly generate the observations used in the Bayesian inference procedure.

\section*{B. Additional Results}
\addcontentsline{toc}{section}{B. Additional Results}

\subsection*{B.1. Stiffness Estimation}

\begin{table}[t]
\centering
\setlength{\tabcolsep}{1.8pt} 
\caption{Stiffness estimation on MIT Fabric Properties dataset (30 objects). ADE is measured in lbf-in$^2$. \textbf{Bold}: best model.}
\label{tab:mit_fabric_stiffness}
\begin{tabular}{lcccc}
\toprule
\textbf{Method} & \textbf{ADE} ($\downarrow$) & \textbf{ALDE} ($\downarrow$) & \textbf{APE} ($\downarrow$) & \textbf{MnRE} ($\uparrow$) \\
\midrule
GPT-4V & 0.563 & 2.380 & 19.986 & 0.210 \\
GPT-5 & 0.126 & 1.053 & 2.887 & 0.452 \\
\midrule
\textbf{Ours} & \textbf{0.040} & \textbf{0.725} & \textbf{1.338} & \textbf{0.553} \\
\bottomrule
\end{tabular}
\end{table}

\begin{figure}[t!]
    \centering
\includegraphics[width=0.5\textwidth]{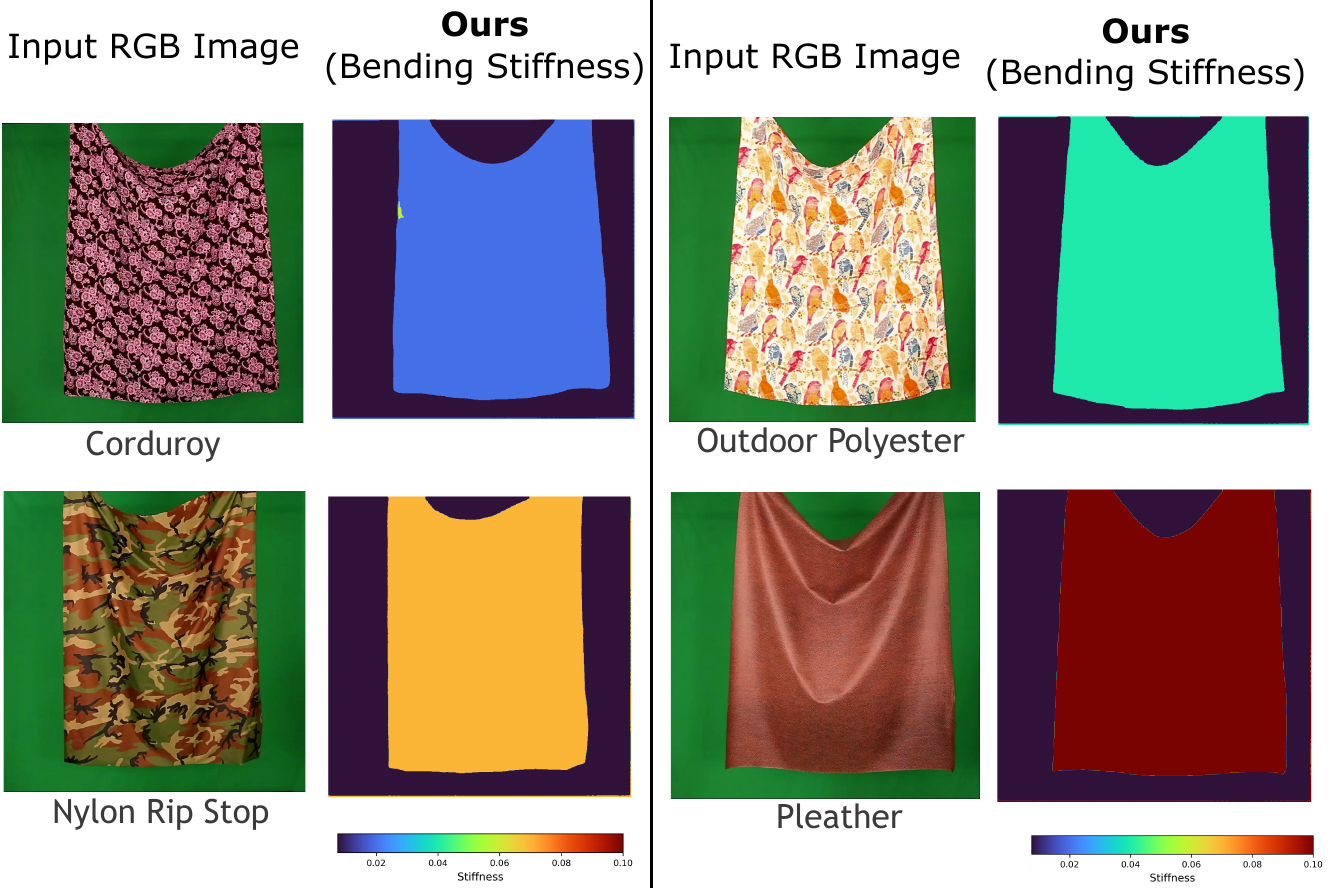}
\caption{Bending stiffness estimation on real fabric samples from MIT Fabric Properties Dataset. Given an input RGB image, \textit{PhysGS} 
produces dense stiffness fields that capture material differences across corduroy, nylon ripstop, 
outdoor polyester, and pleather.}
\vspace{-.3cm}
\label{fig:fabric_viz}
\end{figure}

\vspace{3pt}
\noindent\textbf{Dataset.}
We employ the MIT Fabric Properties Dataset ~\cite{bouman2013estimating} for evaluating mass prediction, 30 different types of real fabric along with measurements of their material properties. Since these are all videos, we curate an image dataset from this, where all the different fabrics are evaluated for their bending stiffness. While these are video datasets, they are captured from a single view, and thus we evaluate our model on one image per fabric. We pick the first frame of every video.

\vspace{3pt}
\noindent\textbf{Metrics.}
We report the same evaluation metrics for bending stiffness estimation (lbf-in$^2$) used in evaluation for mass estimation as above: ADE, ALDE, APE, MnRE.

\vspace{3pt}
\noindent\textbf{Baselines.}
We compare our model against several visual and multimodal baselines on the ABO-500 dataset:
\begin{itemize}
    \item \textbf{GPT-4V}: We provide GPT-4V with the image, and ask it to estimate the physical stiffness of the fabric.
    \item \textbf{GPT-5}: Same prompt as GPT-4V.
\end{itemize}

\vspace{3pt}
\noindent\textbf{Quantitative Results.}
Table~\ref{tab:mit_fabric_stiffness} reports quantitative results comparing our method against GPT-4V and GPT-5 VLM baselines. Across all metrics, \textit{PhysGS} achieves the strongest performance, reducing ADE by 68.3\% compared to GPT-5 and by more than an order of magnitude compared to GPT-4V. Our method also attains the highest MnRE score, indicating substantially improved scale consistency in stiffness estimation. These gains highlight the effectiveness of our Bayesian fusion framework in capturing fine-grained material compliance even in visually ambiguous textile structures.

\noindent\textbf{Qualitative Results.}
Figure~\ref{fig:fabric_viz} presents qualitative bending stiffness estimation results on real fabric samples from the MIT Fabric Properties dataset. The dataset contains diverse materials with visually similar appearances but substantially different mechanical behavior, making stiffness prediction particularly challenging. Across a variety of textile types, including corduroy, nylon ripstop, outdoor polyester, and pleather, \textit{PhysGS} produces dense stiffness fields that clearly delineate material differences. Each predicted stiffness map exhibits smooth spatial variation and preserves mask-level boundaries, reflecting the underlying compliance characteristics of each fabric.

\subsection*{B.2. Terrain Friction Estimation}

\noindent\textbf{Dataset.}
We evaluate terrain friction prediction using the Terrain Class Friction dataset from~\cite{ewen2022these}. The dataset contains paired RGB images and friction measurements for seven common indoor and outdoor terrain classes, including carpet, concrete, laminated flooring, rubber, pebbles, rocks, and wood (see Table \ref{tab:ewen_friction_stats}). Following the protocol in~\cite{ewen2022these}, we assess prediction accuracy against the mean coefficients of friction obtained from their unimodal Gaussian fits for each terrain class.

\begin{table}[h!]
    \centering
    \caption{Mean ($\mu$) and standard deviation ($\sigma$) of coefficients of friction 
    reported in~\cite{ewen2022these} for the Terrain Class Friction Dataset.}
    \label{tab:ewen_friction_stats}
    \begin{tabular}{lcc}
        \toprule
        \textbf{Terrain Class} & $\boldsymbol{\mu}$ & $\boldsymbol{\sigma}$ \\
        \midrule
        Concrete           & 0.543 & 0.065 \\
        Pebbles            & 0.428 & 0.059 \\
        Rocks              & 0.478 & 0.113 \\
        Wood               & 0.372 & 0.055 \\
        Rubber             & 0.616 & 0.048 \\
        Carpet                & 0.583 & 0.068 \\
        Laminated Flooring & 0.311 & 0.045 \\
        \bottomrule
    \end{tabular}
\end{table}

\begin{table}[t]
\centering
\setlength{\tabcolsep}{1.8pt} 
\caption{Friction estimation on Terrain Class Friction dataset (30 objects). ADE is measured in lbf-in$^2$. \textbf{Bold}: best model.}
\label{tab:terrain_class_friction}
\begin{tabular}{lcccc}
\toprule
\textbf{Method} & \textbf{ADE} ($\downarrow$) & \textbf{ALDE} ($\downarrow$) & \textbf{APE} ($\downarrow$) & \textbf{MnRE} ($\uparrow$) \\
\midrule
GPT-4V & 0.129 & 0.315 & 0.286 & 0.747 \\
GPT-5 & 0.146 & 0.253 & 0.291 & 0.779 \\
\midrule
\textbf{Ours} & \textbf{0.126} & \textbf{0.251} & \textbf{0.290} & \textbf{0.783} \\
\bottomrule
\end{tabular}
\end{table}

\begin{figure}[t!]
    \centering
\includegraphics[width=0.5\textwidth]{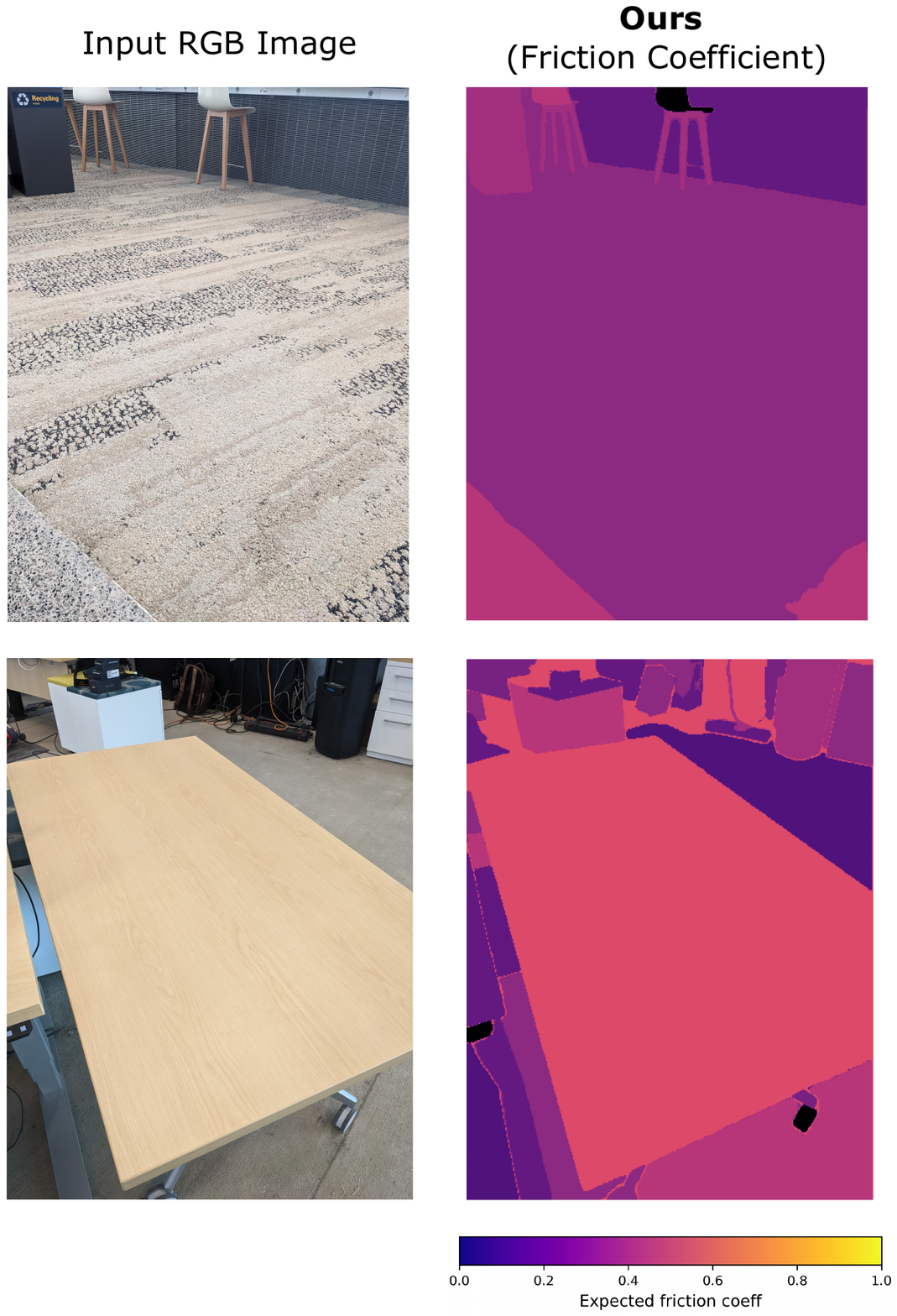}
\caption{Friction estimation on samples from the Terrain Class Friction dataset. \textit{PhysGS} produces smooth, coherent friction maps that differentiate surfaces such as carpet, wood, and composite flooring directly from RGB input.}
\vspace{-.3cm}
\label{fig:terrain_friction_viz}
\end{figure}

\vspace{3pt}
\noindent\textbf{Metrics.}
We report the same evaluation metrics for terrain friction estimation (lbf-in$^2$) used in evaluation for mass estimation as above: ADE, ALDE, APE, MnRE.

\vspace{3pt}
\noindent\textbf{Baselines.}
We compare our model against several visual and multimodal baselines on the ABO-500 dataset:
\begin{itemize}
    \item \textbf{GPT-4V}: We provide GPT-4V with the image, and ask it to estimate the friction of the terrain.
    \item \textbf{GPT-5}: Same prompt as GPT-4V.
\end{itemize}

\vspace{3pt}
\noindent\textbf{Quantitative Results.}
Table~\ref{tab:terrain_class_friction} reports quantitative results comparing our method against GPT-4V and GPT-5 VLM baselines. As the dataset consists of single-object, mostly homogeneous surfaces, the benefits of precise part-level  segmentation are limited in this setting. Nevertheless, our hierarchical prompting scheme enables both global and local reasoning by guiding the VLM to focus on the dominant surface region while still incorporating contextual cues such as reflectance, roughness, and material structure. Across all four metrics, ADE, ALDE, APE, and MnRE, our method performs on par with or better than GPT-4V and GPT-5.

\noindent\textbf{Qualitative Results.} Figure~\ref{fig:terrain_friction_viz} presents qualitative friction estimation results on samples from the Terrain Class Friction dataset. Given an input RGB image, \textit{PhysGS} produces smooth and spatially consistent friction fields that align with the visual regions of each surface. The predicted  maps clearly distinguish materials such as carpet, wood, and composite flooring,  capturing their characteristic friction patterns while preserving coherent region boundaries.

\subsection*{B.3. Outdoor Scene Analysis}

\begin{figure}
    \centering
\includegraphics[width=0.50\textwidth]{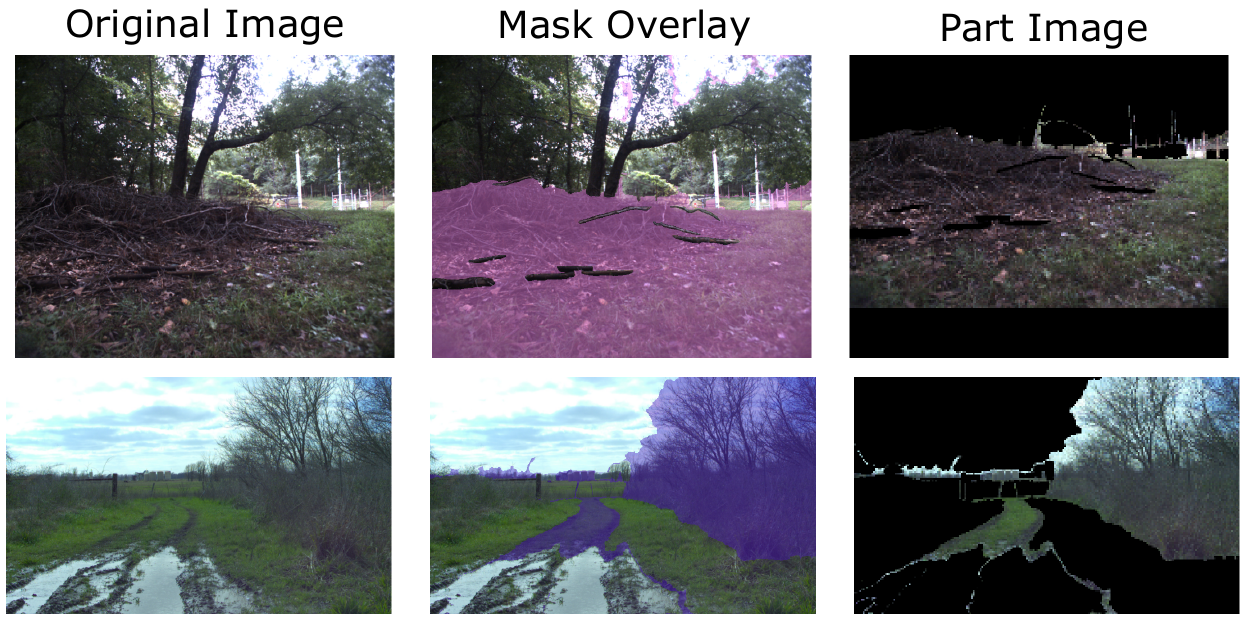}
\vspace{-8.4cm}
\caption{Imprecise masks generated by SAM as can be seen in the mask overlay and the part images. This results in less clear material boundaries and higher downstream uncertainty.}
\vspace{-0.3cm}
\label{fig:bad_sam_masks}
\end{figure}

Figure~\ref{fig:outdoor_scene} shows qualitative results of \textit{PhysGS} applied to outdoor environments with diverse terrain types, vegetation, and natural materials. From a single RGB image, our model predicts material segmentation, friction coefficients, stiffness (Young’s modulus) fields, and total uncertainty (aleatoric + epistemic). These results demonstrate the ability of \textit{PhysGS} to extend beyond controlled indoor settings and operate on unstructured outdoor scenes.

Across all examples, the predicted material maps provide reasonable semantic decomposition of natural surfaces such as gravel, grass, bark, mud, water, and leaf litter. The corresponding friction and stiffness fields reflect meaningful physical differences between these materials: solid regions such as rock, concrete, or bark consistently receive higher stiffness values, whereas deformable surfaces such as mud and grass yield lower estimated moduli. Friction estimates likewise align with expected terrain properties, capturing transitions between slippery, saturated mud and higher-friction vegetation or gravel. 


The total uncertainty maps reveal a strong correlation between uncertainty and the quality of SAM-generated segmentations, consistent with the discussion in the limitations section (see Sec.~\ref{sec:conclusion}). Rows~2 and~3 in Figure~\ref{fig:outdoor_scene} contain dense clutter, irregular textures, or  ambiguous boundaries (e.g., intertwined vegetation or mud–grass transitions), leading SAM to produce noisier part-level masks. As illustrated explicitly in Figure~\ref{fig:bad_sam_masks}, these mask inaccuracies propagate into the part images and result in less reliable material evidence. 
In such cases, \textit{PhysGS} assigns noticeably higher total uncertainty, driven by both epistemic uncertainty from inconsistent material cues and aleatoric uncertainty arising from intra-region variability.

Conversely, rows~1 and~4 in Figure~\ref{fig:outdoor_scene} contain large, spatially coherent surfaces (e.g., gravel, sky, uniform grass), where SAM produces cleaner segmentations. In these settings, \textit{PhysGS} yields lower uncertainty and more stable physical predictions across the scene. Taken together, these results, supported by both Figures~\ref{fig:outdoor_scene} and \ref{fig:bad_sam_masks}, demonstrate that the Bayesian uncertainty estimates are meaningfully sensitive to segmentation quality and reliably signal when the input evidence is less trustworthy.

\section*{C. Additional Experimental Details}
\addcontentsline{toc}{section}{B. Additional Experimental Details}

\begin{figure}
    \centering
\includegraphics[width=0.5\textwidth]{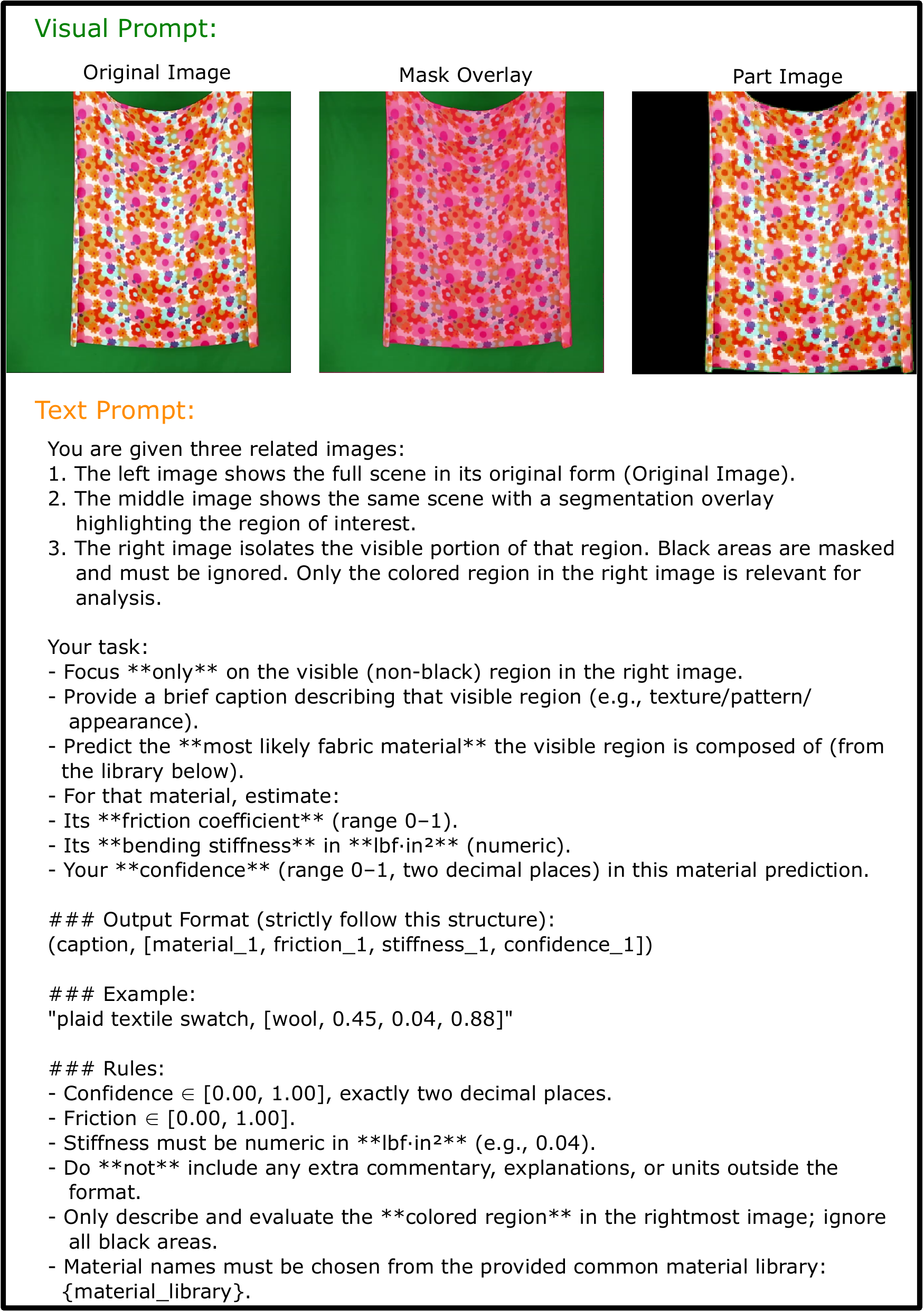}
\caption{VLM Prompt used to obtain material, friction, and bending stiffness predictions for the MIT Fabric Properties dataset.}
\vspace{-0.4cm}
\label{fig:gpt_fabric_prompt}
\end{figure}


\subsection*{C.1. Prompting Details}

\begin{figure}[!t]
    \centering
\includegraphics[width=0.5\textwidth]{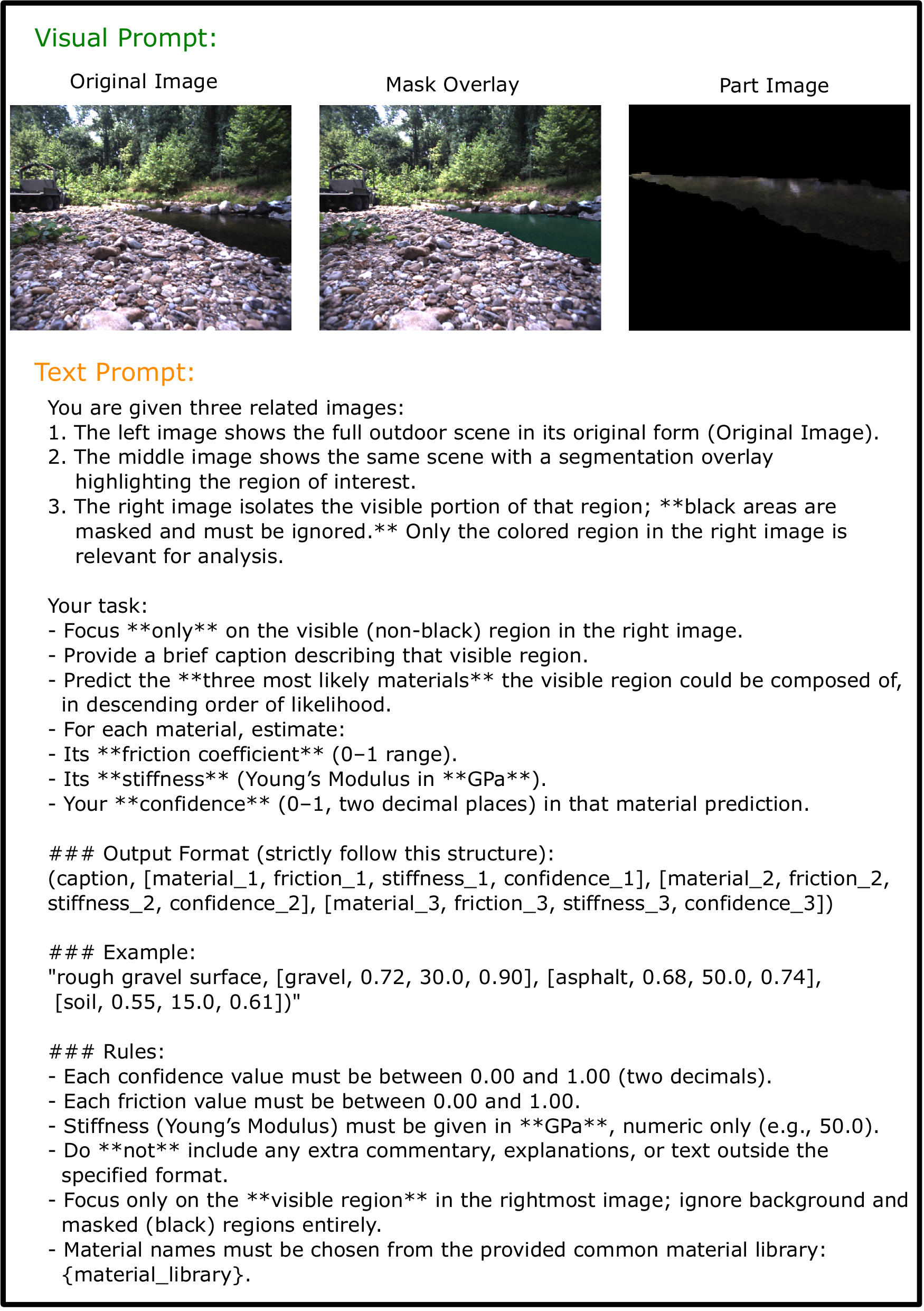}
\caption{VLM prompt used to obtain material, friction, and stiffness predictions for the RUGD dataset. By predicting multiple plausible materials with associated confidences, this prompting strategy enables \textit{PhysGS} to estimate the total uncertainty for each mask.}
\vspace{-.3cm}
\label{fig:outdoor_prompt}
\end{figure}

Figures~\ref{fig:gpt_fabric_prompt} and \ref{fig:outdoor_prompt} show the exact prompting configurations, inspired by~\cite{xu2025gaussianproperty}, used for the MIT Fabric Properties dataset and the RUGD outdoor dataset. In both cases, the VLM is provided with the original RGB image, a segmentation-mask overlay, and an isolated part image. The text prompt directs the model to ignore masked regions and focus only on the visible segment, ensuring that predictions are part-specific rather than influenced by the surrounding scene. We also maintain separate indoor and outdoor material libraries so the VLM selects from the most appropriate set of materials for each environment.

For each part, the VLM returns one or more candidate materials with associated physical properties and confidence scores. Each of these candidate predictions is treated as a confidence-weighted observation within our Bayesian framework, allowing PhysGS to fuse evidence across views and produce consistent material and property estimates. Importantly, the distribution of confidence across multiple materials provides a direct signal of semantic ambiguity. When the VLM is uncertain, often due to noisy or imprecise SAM segmentations, the confidence spread increases, which propagates into higher predictive uncertainty in our property fields, consistent with the trends discussed in the limitations section (Sec. \ref{sec:conclusion}).

\subsection*{C.2. Baseline Details}

\begin{figure}[t]
\vspace*{-10.6cm}
    \centering
\includegraphics[width=0.5\textwidth]{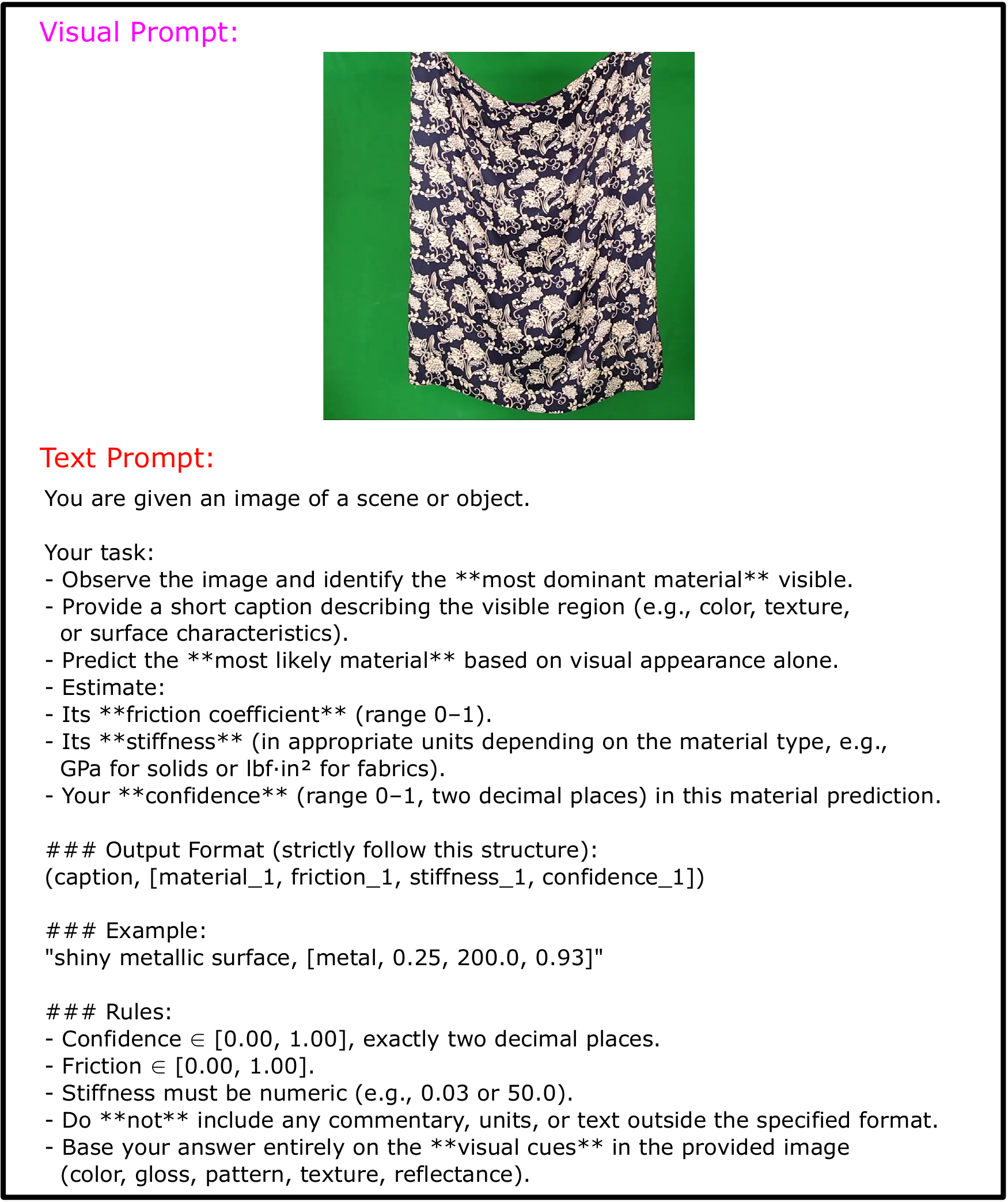}
\caption{Baseline VLM Prompt used to obtain material, friction, and bending stiffness predictions for the MIT Fabric Properties dataset.}
\label{fig:baseline_fabric_prompt}
\end{figure}

To benchmark \textit{PhysGS} against existing vision--language models, we evaluate GPT-4V and GPT-5 on the MIT Fabric Properties and Terrain Class Friction datasets using a simplified prompting strategy tailored for fair comparison (see Figure \ref{fig:baseline_fabric_prompt}). For each image, the VLM receives only the raw RGB frame and is instructed to (1) describe the dominant visible region, (2) predict the most likely material based solely on visual appearance, and (3) estimate a friction coefficient, stiffness value, and confidence score.

This baseline prompt does not include segmentation cues or part-based isolation, and therefore tests each VLM’s ability to infer material and physical properties directly from appearance alone. 
The resulting predictions serve as a reference for evaluating the gains provided by our part-aware prompting, used in \textit{PhysGS}.

\end{document}